\documentclass[runningheads]{llncs}
\usepackage[T1]{fontenc}
\usepackage{graphicx}
\usepackage{cite}
\usepackage{subcaption}
\usepackage{comment}
\usepackage{booktabs}
\usepackage{siunitx}

\usepackage{colortbl}
\usepackage{pgfplots}
\pgfplotsset{compat=newest}

\usepackage{algorithm}
\usepackage{algpseudocode}
\usepackage{amsfonts}
\usepackage{amsmath}
\usepackage{hyperref}

\usepackage{tikz}
\usetikzlibrary{decorations.pathmorphing, positioning, arrows.meta, fit, calc}
\usepackage{graphicx}  % pour \resizebox
\usepackage{mathtools}

\newcommand{\rev}[1]{#1}

\begin{document}
\title{Improving Multivariate Time Series Classification with Class-Wise Training and Model Aggregation}
\titlerunning{Improving MTSC with Class-Wise Training and Model Aggregation}
\author{Mouhamadou Mansour LO\inst{1} \and
Gildas MORVAN\inst{2}\\
Mathieu ROSSI\inst{1} \and Fabrice MORGANTI\inst{1} \and David MERCIER\inst{2}}
\authorrunning{M. LO et al.}
\institute{Univ. Artois, UR 4025, LSEE, F-62400 Béthune, France \and
 Univ. Artois, UR 3926, LGI2A, F-62400 Béthune, France\\
 \email{firstname.lastname@univ-artois.fr}
}
\maketitle              % typeset the header of the contribution
\begin{abstract}
In this paper, we propose a class-wise dimension (channel) selection framework for Multivariate Time Series Classification (MTSC). Rather than applying a single global dimension selection process, the proposed approach independently identifies informative dimensions for each class. A dedicated learning process is subsequently performed for each class, followed by a fusion stage for final prediction. The objective is to improve the generation of discriminative feature representations while reducing the influence of noisy or non-informative dimensions. The proposed framework is evaluated using MiniRocket, a random kernel-based baseline method. Experimental results indicate that class-wise dimension selection improves the quality of extracted representations and can enhance classification performance, particularly in high-dimensional settings. These findings suggest that incorporating class-specific information into the training process represents a promising direction for MTSC, \rev{improving robustness through consistent gains across heterogeneous datasets, and interpretability through the explicit identification of class-relevant dimensions.}

\end{abstract}

\section{Introduction}
\begin{figure}[htbp]
  \centering
  \resizebox{0.8\textwidth}{!}{
    % Prérequis dans le préambule du document principal :
%   \usepackage{tikz}
%   \usetikzlibrary{decorations.pathmorphing, positioning, arrows.meta}

\tikzset{
  %------------------------------------------------------
  % Clé couleur commune à tous les pics
  %------------------------------------------------------
  wavy color/.store in = \wavycolor,
  wavy color = black,
  %------------------------------------------------------
  % Pic « singlestrokes » — Class ?
  %   1 signal par dimension
  %------------------------------------------------------
  singlestrokes/.pic = {
    \foreach \dy/\amp/\seg/\xs/\xe in {
       0.60 / 1.4mm /  8mm / -0.90 / 0.92,
       0.00 / 2.0mm / 10mm / -0.95 / 0.88,
      -0.60 / 1.7mm /  7mm / -0.88 / 0.95%
    }{
      \draw[\wavycolor, line width=1pt, line cap=round,
            decorate,
            decoration={snake, amplitude=\amp, segment length=\seg}]
        (\xs, \dy) -- (\xe, \dy);
    }
  },
  %------------------------------------------------------
  % Pic « multistrokesI » — Class #1
  %   Dim 2 : grandes ondes lentes
  %   Dim 3 : ondes courtes rapides
  %------------------------------------------------------
  multistrokesI/.pic = {
    % -- Dimension 1 (haut) --
    \foreach \dy/\amp/\seg in {
       0.70 / 1.4mm /  9mm,
       0.60 / 1.6mm /  8mm,
       0.50 / 1.3mm /  9mm%
    }{
      \draw[\wavycolor, line width=1pt, line cap=round,
            decorate,
            decoration={snake, amplitude=\amp, segment length=\seg}]
        (-0.92, \dy) -- (0.90, \dy);
    }
    % -- Dimension 2 (milieu) : grandes ondes lentes --
    \foreach \dy/\amp/\seg in {
       0.10 / 2.0mm / 11mm,
       0.00 / 2.3mm / 10mm,
      -0.10 / 1.9mm / 11mm%
    }{
      \draw[\wavycolor, line width=1pt, line cap=round,
            decorate,
            decoration={snake, amplitude=\amp, segment length=\seg}]
        (-0.95, \dy) -- (0.88, \dy);
    }
    % -- Dimension 3 (bas) : ondes courtes rapides --
    \foreach \dy/\amp/\seg in {
      -0.50 / 1.5mm /  7mm,
      -0.60 / 1.8mm /  8mm,
      -0.70 / 1.4mm /  7mm%
    }{
      \draw[\wavycolor, line width=1pt, line cap=round,
            decorate,
            decoration={snake, amplitude=\amp, segment length=\seg}]
        (-0.90, \dy) -- (0.92, \dy);
    }
  },
  %------------------------------------------------------
  % Pic « multistrokesII » — Class #2
  %   Dim 2 : petites ondes rapides  (proche mais distinct)
  %   Dim 3 : grandes ondes lentes   (proche mais distinct)
  %------------------------------------------------------
  multistrokesII/.pic = {
    % -- Dimension 1 (haut) --
    \foreach \dy/\amp/\seg in {
       0.70 / 1.4mm /  9mm,
       0.60 / 1.6mm /  8mm,
       0.50 / 1.3mm /  9mm%
    }{
      \draw[\wavycolor, line width=1pt, line cap=round,
            decorate,
            decoration={snake, amplitude=\amp, segment length=\seg}]
        (-0.92, \dy) -- (0.90, \dy);
    }
    % -- Dimension 2 (milieu) : petites ondes rapides --
    \foreach \dy/\amp/\seg in {
       0.10 / 1.4mm /  8mm,
       0.00 / 1.6mm /  9mm,
      -0.10 / 1.3mm /  8mm%
    }{
      \draw[\wavycolor, line width=1pt, line cap=round,
            decorate,
            decoration={snake, amplitude=\amp, segment length=\seg}]
        (-0.95, \dy) -- (0.88, \dy);
    }
    % -- Dimension 3 (bas) : grandes ondes lentes --
    \foreach \dy/\amp/\seg in {
      -0.50 / 1.9mm / 11mm,
      -0.60 / 2.2mm / 12mm,
      -0.70 / 2.0mm / 11mm%
    }{
      \draw[\wavycolor, line width=1pt, line cap=round,
            decorate,
            decoration={snake, amplitude=\amp, segment length=\seg}]
        (-0.90, \dy) -- (0.92, \dy);
    }
  },
  %------------------------------------------------------
  % Style paramétré pour les boîtes
  %------------------------------------------------------
  classbox/.style args = {#1}{
    rounded corners = 10pt,
    minimum width   = 3.1cm,
    minimum height  = 2.4cm,
    draw = #1!55,
    fill = #1!14,
  },
  %------------------------------------------------------
  % Style des flèches interrogatives (tirets + ?)
  %------------------------------------------------------
  qarrow/.style = {
    ->, #1!75, thick, dashed,
    dash pattern = on 4pt off 2pt,
  },
}

\begin{tikzpicture}[
  font = \sffamily\small\bfseries,
  >   = {Stealth[length=7pt, width=5pt]},
]

  %--- Class #1 : bleu, gauche ---
  \node[classbox=blue] (box1) at (-4.0, 0) {};
  \pic[wavy color=blue!80] at (-4.0, 0) {multistrokesI};
  \node[below=5pt of box1, text=blue] {CLASS \#1};

  %--- Class ? : gris, centre ---
  \node[classbox=gray] (box3) at (0, 0) {};
  \pic[wavy color=gray!65] at (0, 0) {singlestrokes};
  \node[below=5pt of box3] {CLASS ?};

  %--- Class #2 : rouge, droite ---
  \node[classbox=red] (box2) at (4.0, 0) {};
  \pic[wavy color=red!80] at (4.0, 0) {multistrokesII};
  \node[below=5pt of box2, text=red] {CLASS \#2};

  %--- Flèches interrogatives en tirets avec label ? ---
  \draw[qarrow=blue]
    (box3.west) -- node[above=2pt, text=black, font=\normalsize] {?} (box1.east);
  \draw[qarrow=red]
    (box3.east) -- node[above=2pt, text=black, font=\normalsize] {?} (box2.west);

\end{tikzpicture}
  }
  \caption{An Illustrative Three-Dimensional Case Study of an MTSC Problem.}
  \label{fig:MTSC}
\end{figure}
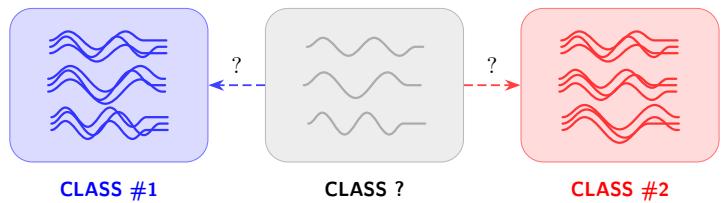

Time Series Classification (TSC) is a supervised learning problem that focuses on assigning predefined class labels to time series instances. The primary objective of TSC is to identify discriminative patterns from labeled training data in order to accurately classify unseen time series sequences. A time series is defined as an ordered sequence of observations collected at regular time intervals, describing the temporal evolution of an event over time. \rev{Formally, a time series is an ordered sequence $\mathbf{X} = (\mathbf{x}_1, \ldots, \mathbf{x}_T)$ of length $T$, where $\mathbf{x}_t \in \mathbb{R}^{d}$ is the observation vector at time step $t$, with $d \geq 2$ in the multivariate case.} In many domains such as autonomous driving, fault diagnosis, patient monitoring, behavioral analysis, or meteorology, the state of a system is often monitored through multiple variables simultaneously measured by different sensors, whose evolution over time forms a Multivariate Time Series. Figure~\ref{fig:MTSC} illustrates a synthetic example of a Multivariate Time Series Classification (MTSC) problem with $d = 3$. A wide range of methods has been proposed in the literature to address TSC problems~\cite{middlehurst_bake_2024}. However, most widely adopted TSC approaches were originally developed for the univariate setting, and their extension to the MTSC problem generally relies on relatively straightforward adaptations of the original frameworks~\cite{ruiz2021great}.

Among the various existing approaches, random convolution kernel-based transformations, including ROCKET~\cite{dempster2020rocket} and its successors MiniRocket~\cite{dempster2021minirocket}, MultiRocket~\cite{tan2022multirocket}, and HYDRA~\cite{dempster2023hydra}, have attracted considerable attention in recent years due to their remarkable trade-off between classification performance and computational efficiency. Deep learning approaches, such as InceptionTime \cite{ismail2020inceptiontime}, are generally natively compatible with Multivariate Time Series. In parallel, hybrid approaches such as HIVE-COTE v2.0, which combine multiple heterogeneous classifiers, typically achieve state-of-the-art accuracy at the expense of a high computational cost~\cite{middlehurst2021hive}. For a comprehensive survey of MTSC methods, the interested reader may refer to~\cite{ruiz2021great}.

The internal dimension-handling mechanisms of most Multivariate methods either rely on random dimension/channel selection (e.g. MiniRocket~\cite{dempster2021minirocket}, or DrCIF~\cite{middlehurst2021hive}), in order to avoid scaling issues as the number of dimensions increases, or exploit all available dimensions simultaneously (e.g., Catch22~\cite{lubbaCatch22CAnonicalTimeseries2019} or WEASEL-MUSE~\cite{schafer2018multivariatetimeseriesclassification}), both of which may lead to suboptimal performance in high-dimensional settings. In the case of ROCKET-based methods, random combinations of dimensions/channels are used during the feature extraction process. In this paper, we investigate an alternative approach that focuses exclusively on dimensions that are relevant to a specific class, with the objective of increasing the probability of generating informative and discriminative dimension combinations for that class. Otherwise, dimensions that are useful for certain classes but irrelevant for others may be combined, potentially leading to information loss.

Our main contribution is the development of a framework that performs dimension selection independently for each class, rather than applying a single global selection process, combined with a class-wise learning strategy followed by a fusion stage for final prediction. We compare the proposed approach against the baseline method and provide a comprehensive analysis of the obtained results.

This article is organized as follows. Section~\ref{sec:relatedworks} first reviews the adaptations of the main random convolution kernel-based methods to the Multivariate setting, as well as the most recent dimensionality reduction algorithms for Multivariate Time Series Classification. Section~\ref{sec:Method} then introduces our proposed framework based on class-wise dimension selection and learning. The performance of its different variants on the UEA Multivariate Time Series Classification Archive \cite{bagnall2018ueamultivariatetimeseries} is analyzed in Section~\ref{sec:Experiments}. Finally, Section~\ref{sec:conclusion} concludes the paper and outlines several future research directions.

\section{Related Work}\label{sec:relatedworks}

\subsection{\rev{Multivariate ROCKET-Based Methods}} 

MTSC methods differ substantially in predictive performance as well as training and inference efficiency. Among them, methods from the ROCKET family have been widely recognized for providing an excellent trade-off between classification accuracy and computational cost.\newline

\textbf{ROCKET} (\textit{Random Convolutional Kernel Transform})~\cite{dempster2020rocket} introduced the use of a large number of randomly generated convolution kernels (typically 10,000) to create activation maps from time series. The resulting activation maps are subsequently summarized using two pooling operators: PPV (Proportion of Positive Values) and GMP (Global Maximum Pooling).\newline 

\textbf{MiniRocket}~\cite{dempster2021minirocket} improves the computational efficiency of ROCKET while maintaining comparable classification performance. This is achieved through a more deterministic design, including a fixed kernel length and a predefined set of 84 kernels with values restricted to $\{-1,2\}$.\newline

\textbf{MultiRocket}~\cite{tan2022multirocket} extends MiniRocket by incorporating first-order differences and additional Pooling Operators extracted from activation maps, namely the Mean of Positive Values (MPV), Mean of Indices of Positive Values (MIPV), and Longest Stretch of Positive Values (LSPV), thereby enriching the feature representation.\newline

\textbf{HYDRA} (\textit{Hybrid Dictionary-ROCKET Architecture})~\cite{dempster2023hydra} combines dictionary-based principles with random convolutional kernels. Kernels are organized into groups, and convolutions are computed for all kernels within each group. At every time step, the kernel yielding the maximum convolution response is selected, enabling the construction of histograms of maximal responses that serve as feature representations for classification. Similar to MultiRocket, first-order differences are incorporated, and the combination of HYDRA and MultiRocket representations further improves classification performance.\newline

Although MiniRocket and related ROCKET-based methods were originally developed for univariate time series, their GitHub implementations generally provide a straightforward multivariate extension based on a Completely-At-Random (CAR) strategy. In MiniRocket, for instance, a random subset of dimensions, with a size ranging from 1 to 9, is selected for each (kernel, dilation, bias) triplet, and the convolution outputs computed independently on these dimensions are summed to produce a single activation map.

While this strategy preserves computational efficiency by preventing complexity from scaling directly with dimensionality, its random selection mechanism may become increasingly suboptimal in high-dimensional settings, reducing the likelihood of generating informative and discriminative dimension combinations.

Consequently, selecting only relevant dimensions through dimensionality reduction may improve the construction of meaningful feature representations while preserving the computational advantages of ROCKET-based methods.

\subsection{\rev{Recent Dimension Reduction Methods for MTSC}}

Dimensionality reduction methods are external preprocessing techniques applied prior to MTSC methods, although many classifiers may already incorporate dedicated mechanisms for handling input dimensions. Their objective is to identify the most relevant dimensions before classification, thereby reducing data dimensionality while preserving discriminative information.\newline

\textbf{ECS} and \textbf{ECP}~\cite{10.1007/978-3-030-91445-5_3,dhariyal2023scalable} are dimensionality reduction methods based on the assumption that useful dimensions should allow different classes to be discriminated through distinct temporal shapes. For each class, a prototype (e.g., Mean, Median, or Median Absolute Deviation) is computed to be its representative time series, and the separability of a dimension is evaluated by measuring the distance between pairs of class prototypes using metrics such as Euclidean distance or Dynamic Time Warping (DTW). The larger the distance, the more distinguishable the two classes are.

ECS evaluates each dimension globally by summing the distances between all pairs of class prototypes and selects dimensions using an elbow-based criterion. 
In contrast, ECP performs pairwise class-based selection and constructs the final subset by taking the union of the dimensions identified as relevant for each class pair.
Consequently, ECS favors dimensions that are globally informative across all classes, whereas ECP preserves dimensions that may only be discriminative for specific classes and therefore only removes dimensions that are non-informative, regardless of the considered classes.\newline

\textbf{\rev{TSelect}}~\cite{Nuyts2025TSelectSR} is a dimensionality reduction method designed to retain the most predictive dimensions while minimizing redundancy. The method operates in two stages. First, the training set is divided into learning and validation subsets. For each dimension, statistical features (minimum, maximum, mean, variance, skewness, and kurtosis) are extracted from time series, and a logistic regression classifier is trained to estimate predictive performance on the validation set using the AUROC metric.

In the second stage, an irrelevant channel selector removes redundant or non-informative dimensions, retaining only complementary dimensions with meaningful predictive value.

\section{Proposed Method}\label{sec:Method}

\subsection{\rev{Framework Overview}}

\rev{
Let $\mathcal{D}=\{(\mathbf{X}_i,y_i)\}_{i=1}^{n}$
denote a multivariate and multiclass time series dataset containing $n$
observations, $d$ dimensions, a sequence length $T$, and $C$ classes, where
$\mathbf{X}_i\in\mathbb{R}^{d\times T}$ and
$y_i\in\{1,\ldots,C\}$.}

\begin{figure}[t]
\centering
  \resizebox{\textwidth}{!}{
    \begin{tikzpicture}[
    font=\scriptsize,
    >=Latex,
    node distance=3mm and 3mm,
    block/.style={
        draw,
        rounded corners=1.5pt,
        align=center,
        minimum height=7mm,
        text width=17mm,
        inner sep=1.5pt,
        line width=0.55pt
    },
    selectionblock/.style={
        draw,
        rounded corners=1.5pt,
        align=center,
        minimum height=7mm,
        text width=18mm,
        inner sep=1.5pt,
        line width=0.55pt
    },
    smallblock/.style={
        draw,
        rounded corners=1.5pt,
        align=center,
        minimum height=7mm,
        text width=14mm,
        inner sep=1.2pt,
        line width=0.55pt
    },
    subset/.style={
        draw,
        rounded corners=1.5pt,
        align=center,
        minimum height=5.5mm,
        minimum width=8mm,
        inner sep=1pt,
        line width=0.55pt
    },
    fusion/.style={
        draw,
        rounded corners=1.5pt,
        align=center,
        minimum height=7mm,
        text width=14mm,
        inner sep=1.2pt,
        line width=0.55pt
    },
    group/.style={
        draw,
        dashed,
        rounded corners=2pt,
        inner xsep=1.3mm,
        inner ysep=2mm,
        line width=0.45pt
    },
    grouplabel/.style={
        font=\tiny,
        align=center
    },
    arrow/.style={
        ->,
        line width=0.65pt
    }
]

% ------------------------------------------------------------------
% Input and class-specific dimension selection
% ------------------------------------------------------------------
\node[block] (data) {
    Training MTS\\
    $\mathcal{D}=\{(\mathbf{X}_i,y_i)\}_{i=1}^{n}$
};

\node[selectionblock, right=2.7mm of data] (selection) {
    Class-wise\\
    dimension selection\\
    (ECS / ECP)
};

\draw[arrow] (data) -- (selection);

% ------------------------------------------------------------------
% Selected subsets
% ------------------------------------------------------------------
\node[subset, right=4mm of selection, yshift=11mm] (s1) {$S_1$};
\node[subset, right=4mm of selection]              (sk) {$S_K$};
\node[subset, right=4mm of selection, yshift=-11mm] (sc) {$S_C$};

\draw[arrow] (selection.east) -- ++(2mm,0) |- (s1.west);
\draw[arrow] (selection.east) -- (sk.west);
\draw[arrow] (selection.east) -- ++(2mm,0) |- (sc.west);

\node at ($(s1)!0.5!(sk)$) {$\vdots$};
\node at ($(sk)!0.5!(sc)$) {$\vdots$};

% ------------------------------------------------------------------
% Class-wise OvR learning
% ------------------------------------------------------------------
\node[smallblock, right=4mm of s1] (f1) {
    $1$ vs Rest\\
    MiniRocket\\[-0.3mm]
    $f_1$
};

\node[smallblock, right=4mm of sk] (fk) {
    $K$ vs Rest\\
    MiniRocket\\[-0.3mm]
    $f_K$
};

\node[smallblock, right=4mm of sc] (fc) {
    $C$ vs Rest\\
    MiniRocket\\[-0.3mm]
    $f_C$
};

\draw[arrow] (s1) -- (f1);
\draw[arrow] (sk) -- (fk);
\draw[arrow] (sc) -- (fc);

% ------------------------------------------------------------------
% Decision scores
% ------------------------------------------------------------------
\node[subset, right=4mm of f1] (z1) {$z_1$};
\node[subset, right=4mm of fk] (zk) {$z_K$};
\node[subset, right=4mm of fc] (zc) {$z_C$};

\draw[arrow] (f1) -- (z1);
\draw[arrow] (fk) -- (zk);
\draw[arrow] (fc) -- (zc);

% ------------------------------------------------------------------
% Score fusion
% ------------------------------------------------------------------
\node[fusion, right=3mm of zk] (fusion) {
    Score fusion\\[0.5mm]
    $\displaystyle \arg\max_{K} z_K$
};

\draw[arrow] (z1.east) -- ++(2mm,0) |- (fusion.west);
\draw[arrow] (zk) -- (fusion);
\draw[arrow] (zc.east) -- ++(2mm,0) |- (fusion.west);

% ------------------------------------------------------------------
% Output
% ------------------------------------------------------------------
\node[block, right=2.7mm of fusion] (output) {
    Predicted class\\
    $\hat{y}$
};

\draw[arrow] (fusion) -- (output);

% ------------------------------------------------------------------
% Compact group boxes and labels
% ------------------------------------------------------------------
\node[group, fit=(s1)(sc)] (groupS) {};
\node[grouplabel, text width=12mm, above=0.7mm of groupS] {Class-specific\\subsets};

\node[group, fit=(f1)(fc)] (groupF) {};
\node[grouplabel, text width=14mm, above=0.7mm of groupF] {Class-wise OvR\\learning};

\node[group, fit=(z1)(zc)] (groupZ) {};
\node[grouplabel, text width=11mm, above=0.7mm of groupZ] {Decision\\scores};

\end{tikzpicture}
  }
\caption{\rev{Overview of the proposed method: for each class $K$, a class-specific dimension subset $S_K$ is selected and a binary one-versus-rest classifier $f_K$ is trained on these selected dimensions only; the $C$ resulting scores are fused to produce the final multiclass prediction.}}
\label{fig:classwise_pipeline}

\end{figure}
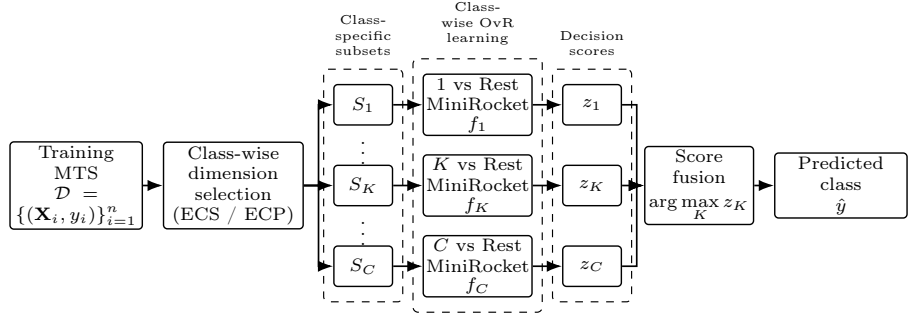

\rev{The proposed framework, illustrated in Figure~\ref{fig:classwise_pipeline}, is motivated by the hypothesis that the dimensions carrying discriminative information may vary across classes and that class-specific combinations of dimensions may provide more informative representations for distinguishing a given class from the others. Rather than relying on a single representation shared across all classes, the framework constructs a dedicated learning branch for each target class $K$.}

\rev{Each branch consists of two successive stages. First, a class-wise dimension
selection procedure identifies a subset
$S_K \subseteq \{1,\ldots,d\}$
containing the dimensions considered informative for distinguishing class
$K$ from the remaining classes. Second, the original multiclass problem is
reformulated as a binary one-versus-rest problem for class $K$, and a
classifier $f_K$ is trained using only the dimensions contained in $S_K$.
Repeating this procedure for all $C$ classes produces the set of
class-specific models
$\{f_1,\ldots,f_C\}$.
At inference time, each model processes the dimensions associated with its
target class, and the resulting decision scores are combined to determine the
final multiclass prediction.}

\rev{The framework therefore differs from conventional global dimension selection
in two respects: the selected subset is allowed to vary across classes, and
each subset is directly associated with a dedicated class-specific learning
problem. The following subsections describe these two stages in detail.}

\subsection{\rev{Class-Wise Dimension Selection}}

\rev{We instantiate the proposed framework using ECS and ECP because their pairwise class-prototype formulation can be naturally conditioned on a target class. Algorithm~\ref{alg:channel_selection} is a class-wise adaptation of the original ECS and ECP methods introduced in~\cite{10.1007/978-3-030-91445-5_3}.}

\rev{For each class $K$ and dimension $j$, let $\mathbf{c}_{K,j}$ denote the
prototype of class $K$ on dimension $j$. The discriminative information
between two classes $K$ and $L$ on dimension $j$ is measured by $\delta_j(K,L)
=
\operatorname{dist}
\left(
\mathbf{c}_{K,j},
\mathbf{c}_{L,j}
\right),
$
where $\operatorname{dist}(\cdot,\cdot)$ denotes the distance measure used by
the corresponding ECS/ECP original configuration.}

\rev{The key difference between the original and proposed formulations lies in the
set of class pairs considered during selection. In the original methods,
dimension relevance is ultimately determined from class pairs covering the
complete multiclass problem, resulting in a single subset $S$ shared by all
classes. In the proposed class-wise formulation, selection for a target class
$K$ considers only the pairs
$
\mathcal{P}_K
=
\{(K,L)\mid L\neq K\}.
$
Consequently, a potentially different subset $S_K$ is obtained for each
class.}

\rev{For the class-wise adaptation of ECS, the relevance score of dimension $j$
with respect to target class $K$ is computed by aggregating only distances
involving $K$:
\begin{equation}
s_j^{(K)}
=
\sum_{L\neq K}
\delta_j(K,L).
\end{equation}
The dimensions are ranked according to $s_j^{(K)}$, and the same elbow-based
selection criterion as in the original ECS method is applied to determine
$S_K$. In contrast, the original ECS formulation aggregates information over
all class pairs and therefore produces a single global subset $S$.}

\rev{For the class-wise adaptation of ECP, dimensions are first ranked separately
for each pair $(K,L)\in\mathcal{P}_K$. Let $S_{K,L}$ denote the subset
selected for the pair $(K,L)$ using the original ECP elbow criterion. The
class-specific subset is then defined as
\begin{equation}
S_K
=
\bigcup_{L\neq K} S_{K,L}.
\end{equation}
The original ECP method instead takes the union of selections obtained from
all class pairs, yielding a single global subset. Restricting this union to
pairs involving $K$ preserves the pairwise principle of ECP while producing
a subset specifically associated with the target class.}
\rev{Algorithm~\ref{alg:channel_selection} summarizes the complete class-wise
selection procedure.}

\begin{algorithm}[ht]
\caption{\rev{Class-Wise Adaptation of ECS and ECP}}
\label{alg:channel_selection}
\renewcommand{\algorithmicrequire}{\textbf{Input:}}
\renewcommand{\algorithmicensure}{\textbf{Output:}}
\begin{algorithmic}[1]

\Require Training dataset $\mathcal{D}$ and selection method
$\mathcal{M}\in\{\mathrm{ECS},\mathrm{ECP}\}$
\Ensure Class-specific dimension subsets
$S=\{S_1,\ldots,S_C\}$

\For{each class $K$ and dimension $j$}
    \State Compute class prototype $\mathbf{c}_{K,j}$
\EndFor

\For{each target class $K$}
    \State $\mathcal{P}_K \gets \{(K,L)\mid L\neq K\}$

    \If{$\mathcal{M}=\mathrm{ECS}$}
        \For{each dimension $j$}
            \State $s_j^{(K)}
            \gets
            \sum_{L\neq K}
            \operatorname{dist}(\mathbf{c}_{K,j},\mathbf{c}_{L,j})$
        \EndFor
        \State Rank dimensions according to $s_j^{(K)}$
        \State Determine the elbow point
        \State $S_K \gets$ dimensions retained by the elbow criterion

    \ElsIf{$\mathcal{M}=\mathrm{ECP}$}
        \State $S_K \gets \emptyset$
        \For{each pair $(K,L)\in\mathcal{P}_K$}
            \For{each dimension $j$}
                \State $\delta_j(K,L)
                \gets
                \operatorname{dist}
                (\mathbf{c}_{K,j},\mathbf{c}_{L,j})$
            \EndFor
            \State Rank dimensions according to $\delta_j(K,L)$
            \State Determine the elbow point
            \State $S_{K,L} \gets$ dimensions retained by the elbow criterion
            \State $S_K \gets S_K \cup S_{K,L}$
        \EndFor
    \EndIf
\EndFor

\State \Return $\{S_1,\ldots,S_C\}$

\end{algorithmic}
\end{algorithm}

\rev{TSelect is not adapted in this study because, unlike ECS and ECP, its
dimension relevance is obtained from the predictive performance of a
classifier trained independently on each dimension. Extending this mechanism
to a class-wise setting would require redefining the relevance estimation for
each target class and would introduce additional classifier-training costs.
We therefore focus on ECS and ECP, whose pairwise formulation allows a direct
class-conditioned adaptation.}

\subsection{\rev{Class-Wise Learning and Score Fusion}}

 Algorithm~\ref{alg:channel_selection} selects, for each class $K\in\{1,\dots,C\}$, a subset of dimensions $S_K\subseteq\{1,\dots,d\}$. This defines the class-specific training subset
\begin{equation}
\mathcal D_K=\left\{(\mathbf{X}_i^{(S_K)}, {y}_i^{(K)})\right\}_{i=1}^n,
\end{equation}
\rev{
where $\mathbf{X}_i^{(S_K)}=(\mathbf{X}_{i,j})_{j\in S_K}$ and
\begin{equation}
y_i^{(K)}=
\begin{cases}
1,& \text{if } y_i=K,\\
0,& \text{otherwise}.
\end{cases}
\end{equation}
For each class $K$, a classifier $f_K$ is trained on $\mathcal{D}_K$. Given a
new observation $\mathbf{X}_p$, the predicted label is
\begin{equation}
\hat{y}_p =\arg\max_{K\in\{1,\dots,C\}} z_K,
\qquad
z_K=\sigma\!\left(f_K(\mathbf{X}_p^{(S_K)})\right),
\end{equation}
}
where $\sigma$ denotes the sigmoid function used to transform decision scores into probability estimates.

\subsection{\rev{Complexity Analysis}}\label{sec:complexity}

\rev{The proposed framework comprises class-wise dimension selection and class-wise MiniRocket transformation. Following Algorithm~\ref{alg:channel_selection}, class-prototype computation (Lines~1--3) costs $O(ndT)$, pairwise class-distance computation (Lines~7--9 and Lines~15--18) costs $O\!\left(C(C-1)dT\right)$, and ranking with elbow detection $O\!\left(Cd\log d\right)$ for ECS (Lines~10--11) and $O\!\left(C^{2}d\log d\right)$ for ECP (Lines~19--20). The resulting selection complexity is therefore
\[
O\!\left(ndT+C(C-1)dT+Cd\log d\right)
\]
for ECS, with the last term replaced by $C^{2}d\log d$ for ECP. The subsequent class-wise MiniRocket transform \cite{dempster2021minirocket} has complexity
\[
O\!\left(k\,n\,T\sum_{c=1}^{C}\min(|S_c|,9)\right),
\]
where $k$ is the total number of MiniRocket kernels and $S_c$ is the subset selected for class $c$. The term $\min(|S_c|,9)$ reflects the MiniRocket implementation constraint that each convolution involves at most nine dimensions. Hence, the computational cost depends on the selected class-specific dimensions rather than on the full dimensionality $d$.}

\section{Results of the Experiments}\label{sec:Experiments}

\subsection{Experimental Settings}
We conducted our experiments on the UEA Multivariate Time Series Classification Archive, which contains 30 multivariate time series datasets. From this archive, we selected 25 datasets with equal-length time series and sequence lengths greater than or equal to 9.

Across the selected datasets, the number of dimensions ranges from 2 to 1345, while sequence lengths vary from 30 to 17,984 time steps. \rev{Following the taxonomy of the UEA Multivariate Time Series Classification Archive~\cite{bagnall2018ueamultivariatetimeseries}, the selected datasets
were grouped according to their application domain into six categories:
Motion Classification (MOTION), Electrocardiography (ECG),
Electroencephalography (EEG), Human Activity
Recognition (HAR), Audio Spectra Classification (AUDIO), and Other miscellaneous categories (OTHER).}

Since multivariate classification methods may exhibit high variability across repeated experiments \cite{ruiz2021great}, each dataset was resampled 30 times, and all methods were evaluated on the resulting folds to ensure robust and reliable performance estimation. For each method and dataset, the standard error across folds was computed to assess performance stability.

In addition, we employed Critical Difference Diagrams (CDD) \cite{demsar2006statistical} to visualize the average rank of methods across datasets, where horizontal cliques indicate the absence of statistically significant differences. We also used Multiple Comparison Matrices (MCM) \cite{ismailfawaz2023approachmultiplecomparisonbenchmark}, in which heatmap colors represent mean performance differences between methods.

\subsection{Baseline Classifier Selection}

\begin{figure}
\scriptsize
\sffamily
\begin{center}

\begin{tabular}{ccccccc}
Mean-Accuracy & \shortstack{Hydra + MultiRocket \\ 73.3494} & \shortstack{MultiRocket \\ 73.1342} & \shortstack{MiniRocket \\ 72.3486} & \shortstack{Hydra \\ 71.4718} & \shortstack{Rocket \\ 71.3443} \\[1ex]
\shortstack{MiniRocket \\ 72.3486} & \cellcolor[rgb]{0.3286,0.4397,0.8696}\shortstack{\rule{0em}{3ex} -1.0008 \\ 12 / 1 / 12 \\ 0.2940} & \cellcolor[rgb]{0.4464,0.5824,0.9574}\shortstack{\rule{0em}{3ex} -0.7856 \\ 12 / 1 / 12 \\ 0.6668} & \cellcolor[rgb]{0.8674,0.8644,0.8626}\shortstack{\rule{0em}{3ex} Mean-Difference \\ r$>$c / r=c / r$<$c \\ Wilcoxon p-value} & \cellcolor[rgb]{0.8771,0.3946,0.3117}\shortstack{\rule{0em}{3ex} 0.8768 \\ 14 / 1 / 10 \\ 0.3327} & \cellcolor[rgb]{0.8204,0.2868,0.2452}\shortstack{\rule{0em}{3ex} 1.0043 \\ 14 / 0 / 11 \\ 0.4842} \\[1ex]
\end{tabular}\\
\begin{tikzpicture}[baseline=(current bounding box.center)]\begin{axis}[hide axis,scale only axis,width=0sp,height=0sp,colorbar horizontal,colorbar style={width=0.25\linewidth,colormap={cm}{rgb255(1)=(83,112,221) rgb255(2)=(220,220,220) rgb255(3)=(209,73,62)},colorbar horizontal,point meta min=-1.21,point meta max=1.21,colorbar/width=1.0em,scaled x ticks=false,xticklabel style={/pgf/number format/fixed,/pgf/number format/precision=3},xlabel={Mean-Difference},}] \addplot[draw=none] {0};\end{axis}\end{tikzpicture}
\end{center}
\caption{Multiple Comparison Matrix for ROCKET-based Methods. Bold values indicate a Wilcoxon p-value $<$ 0.05.}
\label{fig:comp_rocket}
\end{figure}
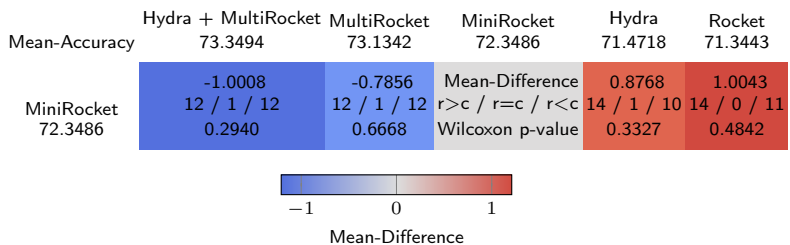

In this study, we compared several methods from the ROCKET family on the 25 selected datasets from the UEA Multivariate Time Series Classification Archive. As shown in Figure~\ref{fig:comp_rocket}, the Multiple Comparison Matrix (MCM) reveals no statistically significant differences among the evaluated methods and variants, including ROCKET, MiniRocket, MultiRocket, HYDRA, and HYDRA + MultiRocket. Notably, the Win/Draw/Loss ratio between MiniRocket and both MultiRocket and HYDRA + MultiRocket is identical (12/1/12).

\rev{Therefore, MiniRocket was selected as the baseline method due to its favorable
trade-off between classification performance and computational efficiency.
This choice is particularly relevant to our framework, which trains one classifier per class and therefore introduces an additional
computational overhead. The complexity analysis provided in
Section~\ref{sec:complexity} further quantifies this additional cost and shows how it
depends on the number of selected dimensions for each class.}

\subsection{Comparison between the Different Variants}

\begin{table}[t]
\caption{\rev{Average classification accuracies with standard errors over 30 resamples for MiniRocket without dimensionality reduction. Dataset abbreviations introduced here are used in the subsequent result tables. Base corresponds to the original setting (all dimensions, no class-wise training), whereas CW denotes the class-wise training variant using all dimensions.}}
\label{tab:accuracies_BCW}
\centering
\scriptsize
\setlength{\tabcolsep}{3pt}

\begin{tabular}{lcc}
\toprule
Datasets (abbreviations) & Base & CW \\
\midrule
ArticularyWordRecognition (AWR) & \textbf{99.39$\pm$0.07} & 99.36$\pm$0.08 \\
AtrialFibrillation (AF)  & \textbf{26.67$\pm$1.83} & 26.22$\pm$1.75 \\
BasicMotions (BM)  & \textbf{100.00$\pm$0.00} & \textbf{100.00$\pm$0.00} \\
Cricket (CR)  & \textbf{99.63$\pm$0.11} & \textbf{99.63$\pm$0.11} \\
DuckDuckGeese (DDG) & 66.53$\pm$0.93 & \textbf{66.93$\pm$1.00} \\
EigenWorms (EW)  & 92.24$\pm$0.43 & \textbf{92.57$\pm$0.54} \\
Epilepsy (EP)  & 99.59$\pm$0.11 & \textbf{99.69$\pm$0.11} \\
EthanolConcentration (EC)  & 47.12$\pm$0.50 & \textbf{49.89$\pm$0.46} \\
ERing (ER)  & 98.07$\pm$0.14 & \textbf{98.25$\pm$0.14} \\
FaceDetection (FD)  & 62.55$\pm$0.17 & \textbf{65.51$\pm$0.17} \\
FingerMovements (FM)  & 57.00$\pm$1.04 & \textbf{58.43$\pm$0.78} \\
HandMovementDirection (HMD) & \textbf{40.09$\pm$0.99} & 39.64$\pm$1.00 \\
Handwriting (HW)  & 55.97$\pm$0.45 & \textbf{56.42$\pm$0.43} \\
Heartbeat (HB)  & 75.82$\pm$0.48 & \textbf{75.97$\pm$0.47} \\
Libras (LIB) & \textbf{92.24$\pm$0.41} & 92.19$\pm$0.41 \\
LSST (LSST) & 64.07$\pm$0.12 & \textbf{64.50$\pm$0.12} \\
MotorImagery  (MI) & 51.77$\pm$0.83 & \textbf{52.70$\pm$0.72} \\
NATOPS (NATO) & 91.04$\pm$0.32 & \textbf{91.43$\pm$0.41} \\
PEMS-SF (PEMS) & 84.26$\pm$0.49 & \textbf{84.59$\pm$0.50} \\
PhonemeSpectra (PS)  & 31.70$\pm$0.13 & \textbf{31.78$\pm$0.13} \\
RacketSports  (RS) & 90.88$\pm$0.31 & \textbf{91.69$\pm$0.35} \\
SelfRegulationSCP1 (SRS1) & 90.14$\pm$0.23 & \textbf{90.31$\pm$0.26} \\
SelfRegulationSCP2 (SRS2) & \textbf{51.89$\pm$0.52} & 51.80$\pm$0.58 \\
StandWalkJump  (SWJ) & \textbf{46.44$\pm$1.79} & \textbf{46.44$\pm$1.92} \\
UWaveGestureLibrary (UW)  & 93.62$\pm$0.20 & \textbf{93.75$\pm$0.19} \\
\bottomrule
\end{tabular}
\end{table}

Table~\ref{tab:accuracies_BCW} reports the classification performance of the original MiniRocket implementation (Base) and the proposed class-wise training variant without dimensionality reduction (CW). Overall, the CW variant consistently achieves slightly better performance, both on average and across individual datasets, with notable improvements on datasets such as EthanolConcentration and FaceDetection, highlighting the benefit of class-specific learning.\newline

\begin{table}[t]
\centering
\scriptsize
\setlength{\tabcolsep}{2.5pt}

\caption{Average classification accuracies with standard errors over 30 resamples for MiniRocket with dimensionality reduction. Base: ECS/ECP global dimensionality reduction without class-wise training; CW + G: ECS/ECP global dimensionality reduction with class-wise training; CW + L: ECS/ECP local dimensionality reduction with class-wise training.}
\label{tab:ecs_ecp_results}

\begin{tabular}{lccc|ccc}
\toprule
& \multicolumn{3}{c}{ECS} & \multicolumn{3}{c}{ECP} \\
\cmidrule(lr){2-4} \cmidrule(lr){5-7}
Dataset & Base & CW+G & CW+L & Base & CW+G & CW+L \\
\midrule
AWR  & 98.40$\pm$0.18 & 98.31$\pm$0.21 & \textbf{98.47$\pm$0.13} & 99.31$\pm$0.07 & 99.31$\pm$0.07 & \textbf{99.34$\pm$0.07} \\
AF   & \textbf{27.33$\pm$1.97} & 27.11$\pm$1.80 & 26.67$\pm$2.08 & 26.44$\pm$1.85 & \textbf{27.56$\pm$1.80} & \textbf{27.56$\pm$1.88} \\
BM   & 99.92$\pm$0.08 & \textbf{100.00$\pm$0.00} & \textbf{100.00$\pm$0.00} & \textbf{100.00$\pm$0.00} & \textbf{100.00$\pm$0.00} & \textbf{100.00$\pm$0.00} \\
CR   & 98.89$\pm$0.23 & 98.89$\pm$0.22 & \textbf{99.21$\pm$0.14} & 99.63$\pm$0.11 & \textbf{99.68$\pm$0.11} & \textbf{99.68$\pm$0.11} \\
DDG  & \textbf{69.33$\pm$1.17} & 68.93$\pm$1.20 & 68.13$\pm$0.98 & \textbf{69.13$\pm$0.84} & 68.67$\pm$1.10 & 68.20$\pm$0.97 \\
EW   & 83.97$\pm$0.88 & 84.55$\pm$0.87 & \textbf{84.61$\pm$0.80} & 86.95$\pm$0.61 & \textbf{87.05$\pm$0.69} & 86.90$\pm$0.69 \\
EP   & 96.59$\pm$0.34 & 96.69$\pm$0.34 & \textbf{97.39$\pm$0.30} & \textbf{99.15$\pm$0.15} & 99.11$\pm$0.16 & 98.67$\pm$0.17 \\
EC   & 47.83$\pm$0.45 & 51.04$\pm$0.51 & \textbf{51.62$\pm$0.55} & 47.00$\pm$0.51 & 50.08$\pm$0.44 & \textbf{51.34$\pm$0.52} \\
ER   & 87.72$\pm$1.17 & 87.56$\pm$1.17 & \textbf{96.09$\pm$0.27} & \textbf{98.12$\pm$0.16} & 98.10$\pm$0.12 & 98.00$\pm$0.17 \\
FD   & 61.99$\pm$0.18 & 61.95$\pm$0.17 & \textbf{64.28$\pm$0.20} & 61.04$\pm$0.19 & 61.10$\pm$0.18 & \textbf{63.38$\pm$0.16} \\
FM   & 54.40$\pm$0.75 & 53.97$\pm$0.97 & \textbf{55.33$\pm$0.92} & 54.07$\pm$1.04 & \textbf{55.90$\pm$0.75} & 55.10$\pm$0.72 \\
HMD  & 41.04$\pm$0.92 & 40.05$\pm$0.89 & \textbf{44.50$\pm$1.02} & 40.50$\pm$0.89 & 38.78$\pm$0.83 & \textbf{42.03$\pm$1.03} \\
HW   & 38.92$\pm$0.94 & 38.75$\pm$0.90 & \textbf{40.11$\pm$0.53} & 56.04$\pm$0.41 & \textbf{56.13$\pm$0.42} & 56.09$\pm$0.41 \\
HB   & 74.62$\pm$0.47 & 74.65$\pm$0.45 & \textbf{74.98$\pm$0.41} & 74.55$\pm$0.44 & 74.85$\pm$0.49 & \textbf{74.93$\pm$0.41} \\
LIB  & \textbf{86.94$\pm$1.09} & 86.78$\pm$1.15 & 86.56$\pm$0.75 & 92.11$\pm$0.42 & 92.02$\pm$0.49 & \textbf{92.19$\pm$0.43} \\
LSST & \textbf{54.57$\pm$0.99} & 54.38$\pm$1.00 & 54.46$\pm$0.92 & 63.84$\pm$0.12 & 64.03$\pm$0.12 & \textbf{64.43$\pm$0.12} \\
MI   & \textbf{51.03$\pm$0.78} & 50.47$\pm$0.71 & 49.93$\pm$0.67 & 50.47$\pm$0.75 & \textbf{50.63$\pm$0.83} & 49.67$\pm$0.75 \\
NATO & 85.91$\pm$0.36 & 85.91$\pm$0.40 & \textbf{86.15$\pm$0.39} & 90.41$\pm$0.32 & 90.44$\pm$0.37 & \textbf{91.00$\pm$0.43} \\
PEMS & \textbf{85.39$\pm$0.51} & 84.62$\pm$0.45 & 85.20$\pm$0.46 & 84.45$\pm$0.44 & \textbf{85.38$\pm$0.48} & 84.93$\pm$0.47 \\
PS   & 31.78$\pm$0.11 & \textbf{31.81$\pm$0.14} & \textbf{31.81$\pm$0.11} & \textbf{31.86$\pm$0.11} & 31.75$\pm$0.10 & 31.68$\pm$0.11 \\
RS   & 88.29$\pm$0.35 & 87.98$\pm$0.34 & \textbf{88.84$\pm$0.37} & 88.79$\pm$0.41 & 88.84$\pm$0.36 & \textbf{88.90$\pm$0.43} \\
SRS1 & \textbf{89.34$\pm$0.25} & 89.17$\pm$0.28 & 89.43$\pm$0.26 & 89.06$\pm$0.27 & \textbf{89.27$\pm$0.26} & 88.95$\pm$0.27 \\
SRS2 & 51.41$\pm$0.54 & 52.02$\pm$0.51 & \textbf{52.31$\pm$0.52} & 51.69$\pm$0.71 & \textbf{51.89$\pm$0.48} & \textbf{51.89$\pm$0.60} \\
SWJ  & 43.33$\pm$2.19 & \textbf{43.56$\pm$1.85} & 42.00$\pm$1.78 & \textbf{44.44$\pm$1.92} & \textbf{44.44$\pm$1.95} & \textbf{44.44$\pm$1.87} \\
UW   & 70.33$\pm$0.46 & 70.44$\pm$0.43 & \textbf{82.94$\pm$0.66} & 93.77$\pm$0.18 & 93.75$\pm$0.22 & \textbf{93.93$\pm$0.24} \\
\bottomrule
\end{tabular}
\end{table}

Table~\ref{tab:ecs_ecp_results} presents the results obtained with the Base model, CW + G (class-wise training with global dimension selection), and CW + L (class-wise training with local dimension selection) using ECS and ECP dimensionality reduction. While CW + G mainly reflects the contribution of class-wise training alone, CW + L highlights the additional benefits of combining class-wise learning with local dimension selection. Overall, both CW + L variants outperform their respective baselines (ECS Base and ECP Base) on most datasets, emphasizing the effectiveness of local dimension selection.

\begin{figure}[!h]
     \begin{center}
             \includegraphics[width=0.8\textwidth]{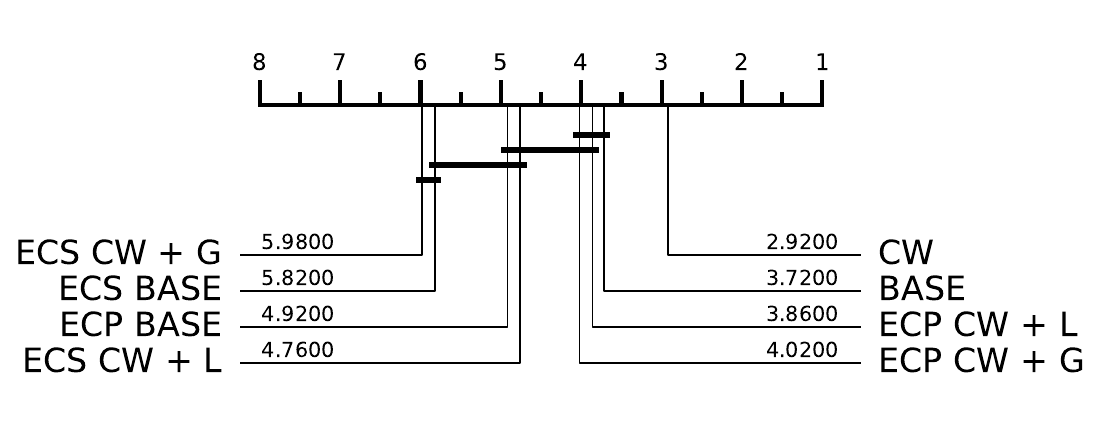}
     \end{center}
    \caption{Critical Difference Diagram of the different MiniRocket variants.}
    \label{fig:CDD}
\end{figure}

Figure~\ref{fig:CDD} compares the developed MiniRocket variants using a Critical Difference Diagram. The CW variant without dimensionality reduction achieves the best average rank and significantly outperforms the second-ranked method, namely the original baseline (Base), demonstrating strong robustness across datasets.

The ECP CW + L variant achieves a ranking comparable to the original baseline while exhibiting superior performance on specific dataset categories (e.g., OTHER). Similarly, ECS CW + L attains a better rank than both ECS Base and ECP Base, further highlighting the benefit of integrating class-wise learning with local dimension selection.\newline

\begin{table}[ht]
\centering
\caption{Comparison by problem category}
\scriptsize
\setlength{\tabcolsep}{4pt}

\resizebox{0.8\linewidth}{!}{
\begin{tabular}{lcccccc}
\hline
Model & ECG & EEG & MOTION & HAR & AUDIO & OTHER \\
\hline
BASE    & \textbf{36.55} & 58.90 & 95.81 & 91.22 & 58.02 & 65.14 \\
CW      & 36.33 & \textbf{59.73} & \textbf{95.96} & \textbf{91.44} & \textbf{58.22} & \textbf{66.32} \\
\hline
ECS BASE & \textbf{35.33} & 58.20 & 91.18 & 83.72 & \textbf{58.57} & 62.59 \\
ECS CW + L  & 34.33 & \textbf{59.29} & \textbf{91.53} & \textbf{86.36} & 58.30 & \textbf{63.76} \\
\hline
ECP BASE & 35.44 & 57.80 & \textbf{93.12} & 90.89 & \textbf{58.51} & 65.09\\
ECP CW + L  & \textbf{36.0} & \textbf{58.50} & \textbf{93.12} & \textbf{90.93} & 58.26 &  \textbf{66.90}\\
\hline
\end{tabular}
}
\label{tab:compperfbyPC}
\end{table}

Table~\ref{tab:compperfbyPC} compares the performance of Base, CW, and CW + L across dataset categories under three configurations: without dimensionality reduction, with ECS, and with ECP. The results indicate that the EEG and OTHER categories particularly benefit from the class-wise strategy. Moreover, ECP CW + L surpasses the original MiniRocket on OTHER and AUDIO datasets, whereas ECS CW + L improves performance on EEG and AUDIO datasets.

\begin{figure}[h]
    \centering

    \begin{subfigure}{0.45\textwidth}
        \centering
        \includegraphics[width=\linewidth]{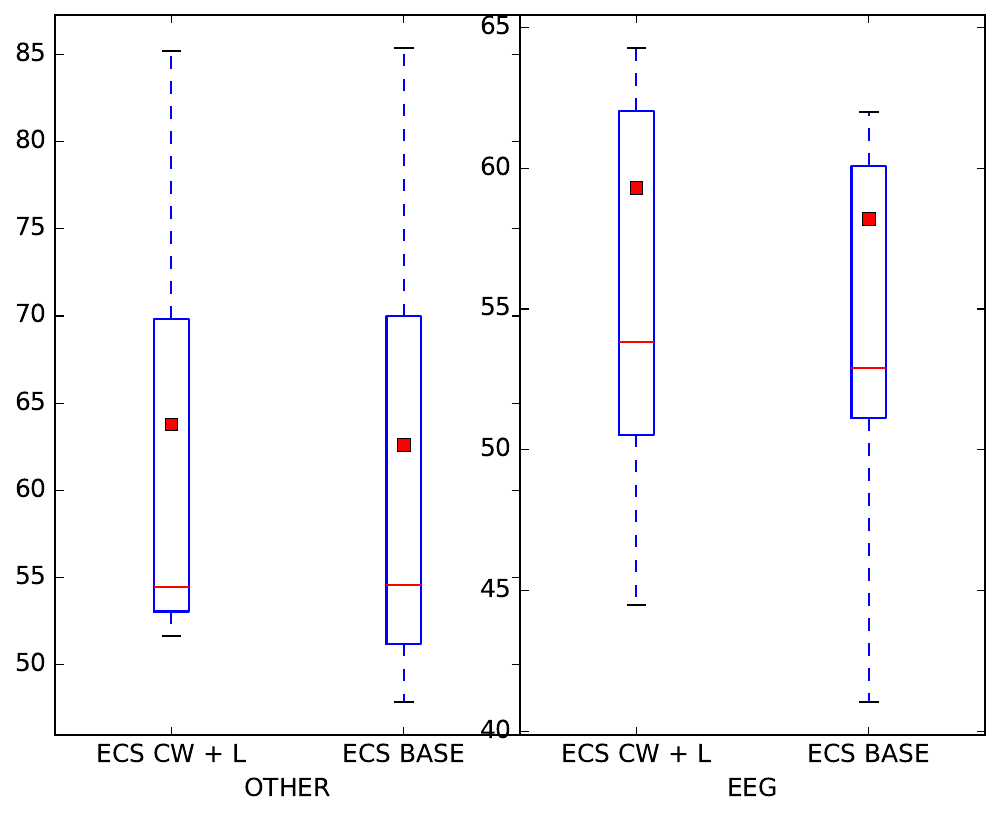}
        \caption{Performances}
    \end{subfigure}
    \hfill
    \begin{subfigure}{0.45\textwidth}
        \centering
        \includegraphics[width=\linewidth]{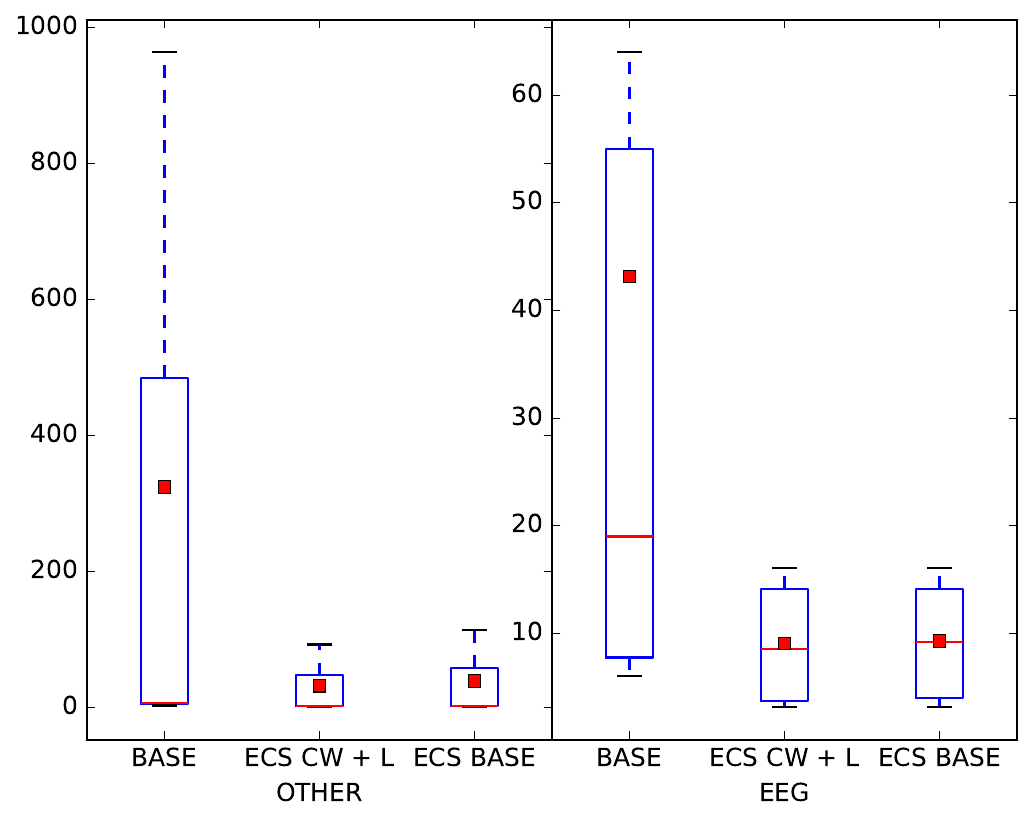}
        \caption{Number of selected dimensions}
    \end{subfigure}

    \caption{Comparison of classification performances and mean number of selected dimensions per classifier between ECS~Base and ECS~CW~+~L across the OTHER and EEG dataset categories.}
    \label{fig:perfECSOTEEG}
\end{figure}

\begin{figure}[h]
    \centering

    % Première ligne
    \begin{subfigure}{0.45\textwidth}
        \centering
        \includegraphics[width=\linewidth]{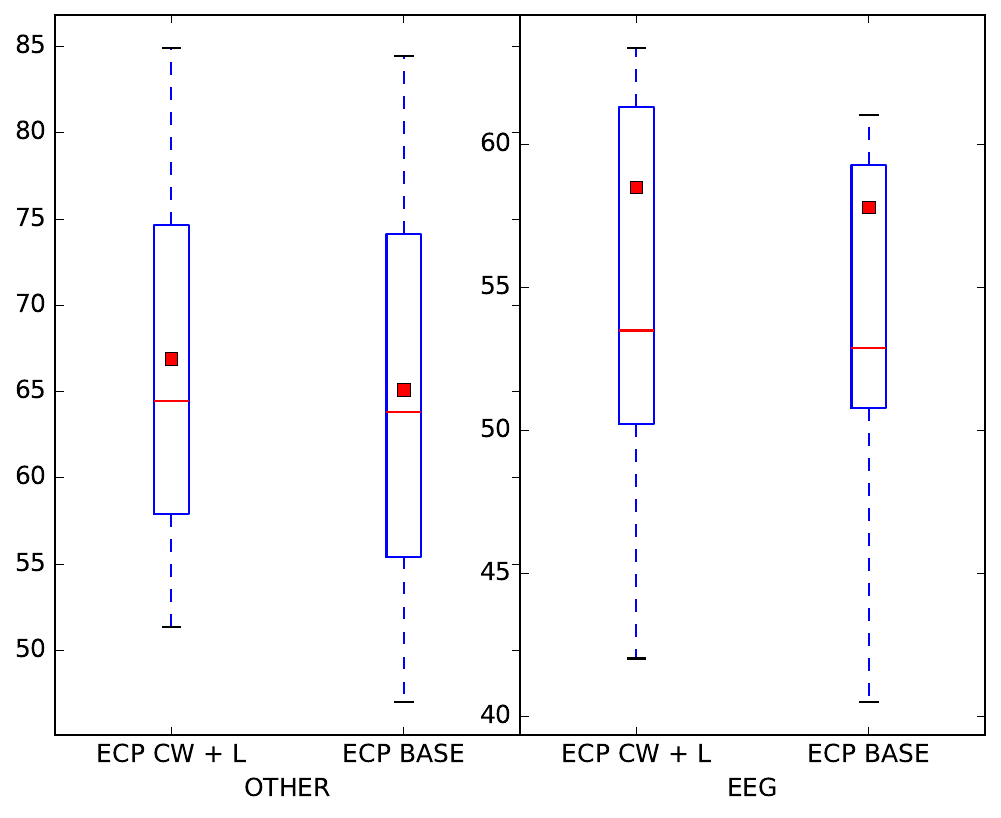}
        \caption{Performances}
    \end{subfigure}
    \hfill
    \begin{subfigure}{0.45\textwidth}
        \centering
        \includegraphics[width=\linewidth]{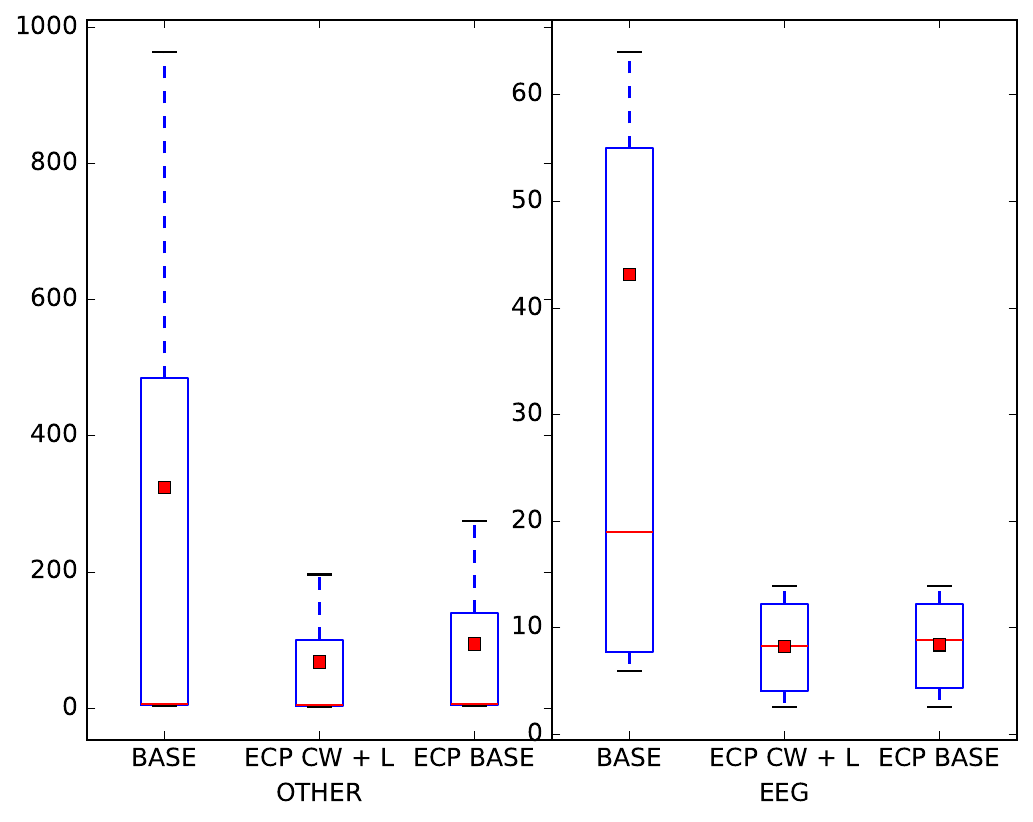}
        \caption{Number of selected dimensions}
    \end{subfigure}

    \caption{Comparison of classification performances and mean number of selected dimensions per classifier between ECP~Base and ECP~CW~+~L across the OTHER and EEG dataset categories.}
    \label{fig:perfECPOTEEG}
\end{figure}

Overall, ECS and ECP combined with the CW + L framework consistently outperform their corresponding baseline variants across nearly all dataset categories, supporting the effectiveness of combining class-wise learning with local dimensionality reduction (Figures~\ref{fig:perfECSOTEEG} and~\ref{fig:perfECPOTEEG}).

\begin{table}[ht]
\centering
\caption{\rev{Comparison by number of dimensions and number of classes}}
\scriptsize
\setlength{\tabcolsep}{4pt}
\resizebox{\linewidth}{!}{
\begin{tabular}{lccccc|cccc}
\toprule
& \multicolumn{5}{c|}{\# Dimensions} & \multicolumn{4}{c}{\# Classes} \\
\cmidrule(lr){2-6} \cmidrule(lr){7-10}
Model & 2 & 3--4 & 5--9 & 10--49 & 50+ & 2 & 3--4 & 5--9 & 10+ \\
\midrule
BASE       & \textbf{59.45} & 73.47 & 86.02 & 54.95 & 68.18 & 64.86 & 64.39 & 87.62 & 73.83 \\
CW         & 59.20 & \textbf{74.07} & \textbf{86.23} & \textbf{55.32} & \textbf{69.13} & \textbf{65.78} & \textbf{64.79} & \textbf{87.91} & \textbf{73.97} \\
\midrule
ECS BASE   & \textbf{57.13} & 64.12 & 83.09 & 53.28 & 68.47 & 63.79 & 63.47 & 80.44 & 68.25 \\
ECS CW + L & 56.61 & \textbf{68.35} & \textbf{83.41} & \textbf{54.44} & \textbf{68.50} & \textbf{64.37} & \textbf{64.43} & \textbf{83.85} & \textbf{68.43} \\
\midrule
ECP BASE   & 59.27 & 73.08 & 84.90 & 54.20 & 67.92 & 63.47 & 63.76 & 87.13 & 73.79 \\
ECP CW + L & \textbf{59.80} & \textbf{73.74} & \textbf{85.01} & \textbf{54.95} & \textbf{68.22} & \textbf{63.98} & \textbf{64.70} & \textbf{87.15} & \textbf{73.90} \\
\bottomrule
\end{tabular}
}
\label{tab:compperfbyNDNC}
\end{table}

\rev{Table~\ref{tab:compperfbyNDNC} compares performance according to the number of dimensions and the number of classes.} The CW variant without dimensionality reduction outperforms the original MiniRocket on datasets containing at most four classes ($\leq 4$) and at least three dimensions ($\geq 3$), while achieving comparable performance elsewhere.

Regarding dimensionality, ECP CW + L performs better than the baseline on low-dimensional datasets ($\leq 4$) and remains competitive on higher-dimensional datasets ($\geq 10$). In contrast, ECS CW + L achieves superior performance on high-dimensional datasets ($\geq 50$) compared to the baseline.\newline

\begin{table}[ht]
\centering
\caption{Comparison of average number of selected dimensions by problem category}

\scriptsize
\setlength{\tabcolsep}{4pt}

\resizebox{0.8\linewidth}{!}{
\begin{tabular}{lcccccc}
\hline
Model & ECG & EEG & MOTION & HAR & AUDIO & OTHER \\
\hline
BASE    & 3.0 & 43.16 & 7.5 & 6.33 & 472.33 & 324.0 \\
\hline
ECS Global & 1.53 & 9.28 & \textbf{2.3} & 2.87 & 111.53 & 39.25 \\
ECS Local  & \textbf{1.47} & \textbf{9.06} & 2.77 &  \textbf{2.71} &  \textbf{96.35} &  \textbf{31.99} \\
\hline
ECP Global & 1.98 & 8.42 & 6.2 & 4.92 & 170.84 & 94.57\\
ECP Local  &  \textbf{1.78} &  \textbf{8.24} &   \textbf{6.04} &  \textbf{4.30} &  \textbf{125.20} &   \textbf{67.97}\\
\hline
\end{tabular}
}
\label{tab:compMNSDbyPC}
\end{table}

Table~\ref{tab:compMNSDbyPC} reports the average number of selected dimensions per classifier across dataset categories. While Base approaches train a single classifier, CW-based approaches train one classifier per class. ECS and ECP Base already provide substantial dimensionality reduction, with ECS selecting markedly fewer dimensions than ECP. 

Furthermore, local selection substantially reduces the number of dimensions per classifier for the OTHER and AUDIO categories in both ECS and ECP. Interestingly, the impact of this reduction varies across categories: in AUDIO, fewer selected dimensions are associated with lower performance, whereas in OTHER, dimensionality reduction improves classification accuracy. These findings suggest that the effectiveness of local dimensionality reduction strongly depends on the problem category and is in most cases beneficial.

\section{Conclusion and Prospects}\label{sec:conclusion}

\rev{In this study, we introduced a class-wise framework for multivariate time series classification that identifies the most relevant dimensions for each class, trains a dedicated classifier on each resulting subset, and produces the final prediction through classifier fusion. The results demonstrate that such a strategy  provides consistent improvements in classification performance across heterogeneous datasets (the class-wise variant achieves the best average rank and significantly outperforms the original MiniRocket), with gains depending on the problem category, and that it is particularly beneficial for methods relying on random dimension selection.} \rev{Moreover, the selected class-specific subsets enhance interpretability by explicitly indicating which dimensions distinguish each class from the others (e.g., which sensors characterize a given activity in HAR problems).}

Future work may extend the proposed framework to other ROCKET-based methods as well as additional approaches from the time series classification literature. Another promising direction would be the adaptation of ChannelScorer-based dimensionality reduction methods to support local dimension selection.

\begin{credits}
\subsubsection{\ackname} \rev{The authors thank the anonymous reviewers for their careful reading of this paper and their constructive comments, which helped improve the manuscript.} This research project, supported and financed by the French ANR (Agence Nationale pour la Recherche), is part of the Labcom (Laboratoire Commun) MYEL (MobilitY and Reliability of Electrical chain Lab) involving LSEE, LGI2A and CRITTM2A (ANR-22-LCV2-0001 MYEL).

\subsubsection{\discintname}
The authors have no competing interests to declare that are
relevant to the content of this article.
\end{credits}

\bibliographystyle{splncs04}
\bibliography{bibliography}

\end{document}